# Control of synaptic plasticity in neural networks *

*Mohammad Modiri* **

## Abstract

The brain is a nonlinear and highly Recurrent Neural Network (RNN). This RNN is surprisingly plastic and supports our astonishing ability to learn and execute complex tasks. However, learning is incredibly complicated due to the brain's nonlinear nature and the obscurity of mechanisms for determining the contribution of each synapse to the output error. This issue is known as the Credit Assignment Problem (CAP) and is a fundamental challenge in neuroscience and Artificial Intelligence (AI). Nevertheless, in the current understanding of cognitive neuroscience, it is widely accepted that a feedback loop systems play an essential role in synaptic plasticity. With this as inspiration, we propose a computational model by combining Neural Networks (NN) and nonlinear optimal control theory. The proposed framework involves a new NN-based actor–critic method which is used to simulate the error feedback loop systems and projections on the NN's synaptic plasticity so as to ensure that the output error is minimized.

*Keywords:* learning rule, neural network, nonlinear control, reservoir computing.

---

* This article is part of a new learning rule that shared to get feedback from researchers.

** Email address: Mohammad.Modiri@Gmail.com

# 1. Proposed framework

In this section, we propose a novel brain-inspired learning rule. The main idea is to use a RL-based method to control synaptic plasticity. Fig. 1 illustrates the main ideas underlying this computational model that applied in a time series classification task.

In the proposed framework, synaptic plasticity reformulated as an optimal tracking problem. For this purpose, the one-hot output vector representing the true class of each time series pattern is imitated, similar to the length of each time series pattern, and is considered the reference trajectory as illustrated in Fig. 1 (C) bottom subplot. Since the reservoir is a special class of high-dimensional nonlinear dynamic systems, where both state and time are continuous. Therefore, continuous-time formulation of the HJB equation and ADP can be applied to derive the learning rule (optimal control law). The derived learning rule forces the proposed framework's output (predicted class in classification tasks) to mimic the reference trajectory (true class in classification tasks) that implicitly minimized the cross-entropy. The following subsections elaborate on the network architecture, and the learning rule in further detail.

## *1.1 Network architecture and dynamics*

The proposed framework consists of a feedforward and feedback pathway. The main parts of the feedforward module are:

- Input data encoding layer.
- Recurrent neural network (RNN).
- Linear classifier (output function).

The feedback module is composed of:

- Critic RC.
- Actor RC.

Fig. 1 shows the block diagram of the proposed framework architecture. The functionality of the proposed architecture is based on the following steps:

a. The encoding layer projects input patterns $x(t)$ into the RNN. In spiking neural networks, each real-value of a data vector is converted into discrete spike trains, suitable for processing in the spike network as spike neural networks can only process discrete spike trains.

b. In the pertaining process, a supervised learning algorithm like Gradient descent, Ridge, etc., is employed to adjust the decoder parameters associated with each training sample. Once the decoder is trained, it is considered constant during the reservoir training process.

c. In the RNN training process. An RL-based algorithm proposed in subsection 1.2 is applied to make the RNN learn temporal relations between the input patterns $x(t)$ and desired output $y(t)$ by adjusting the parameters in the RNN using a novel ADP approach. In the proposed ADP mechanism, two RC-based actor and critic are utilized to approximate true reservoir parameters.

d. Model recall on new data.

The feedforward pathway resembles the structure of traditional neural networks, however the methodology of the feedback pathway especially proposed in this study differs significantly from classical artificial intelligence approaches.

**Remark 1** For the sake of brevity, time dependence is suppressed while denoting variables of dynamical systems. For instance, the notations $x(t)$, $y(t)$, $u(t)$, $v(t)$ and $e(t)$ are rewritten as $x$, $y$, $u$, $v$ and $e$.

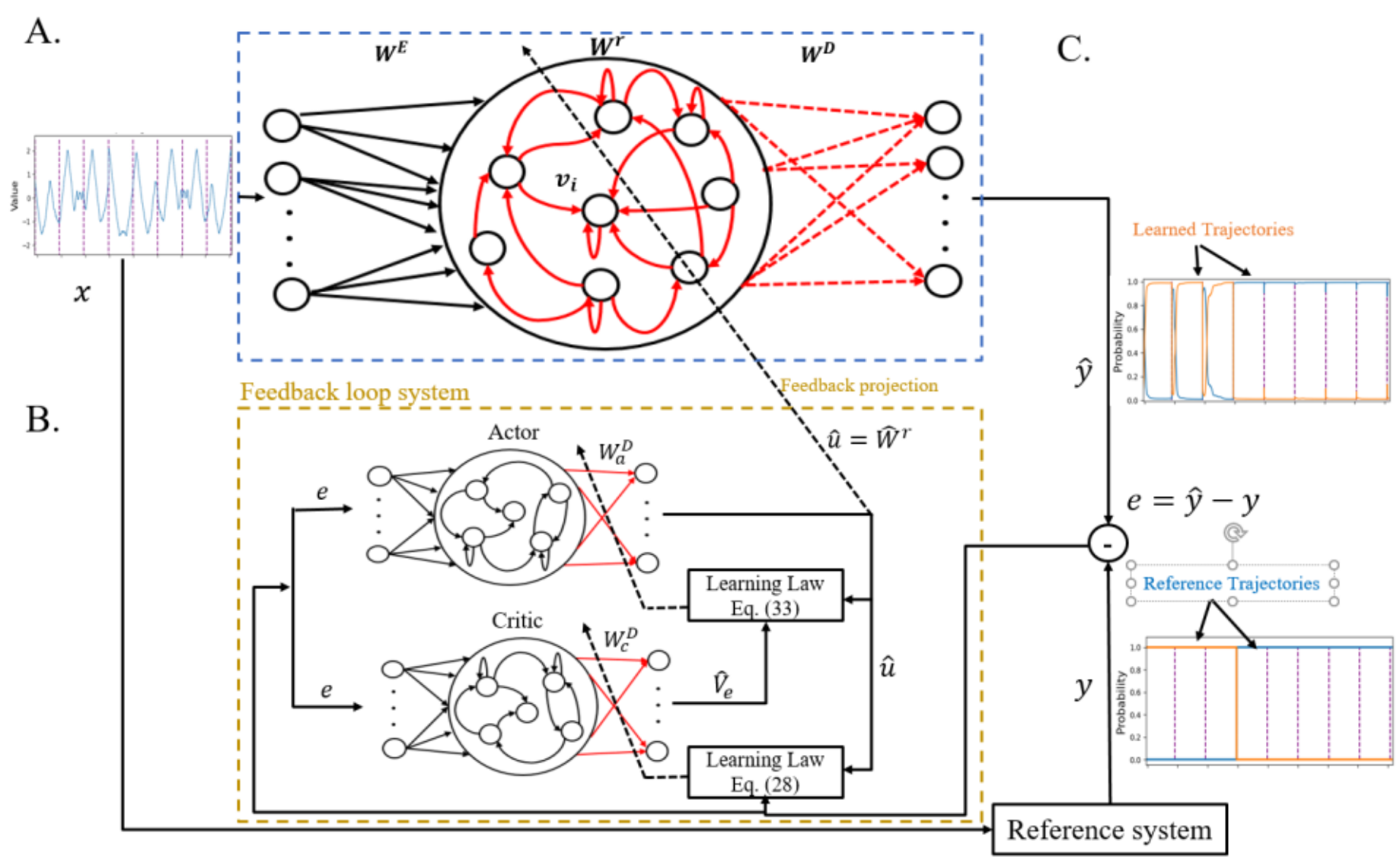

**Fig. 1.** A schematic of the proposed architecture (plastic weights in red). (A) The feedforward pathway consists of an encoding layer with fix and random parameters, a RNN with trainable parameter and a decoder layer, which is trained in the pre-training process and stay constant during the RNN training process. (B) In the feedback pathway, the nonlinear optimal control is applied for estimating RNN parameters. It consists of actor and critic neural networks. (C) The bottom subplot shows the reference trajectories, and the top subplot shows the learned trajectories.

As illustrated in Fig. 1, The central element of the proposed framework is a RNN, i.e., a recurrent network with a structure denoted by the adjacency matrix $W^r \in \mathbb{R}^{n^r \times n^r}$. Here, the RNN consists of $n^r$ neurons, for which the membrane potential dynamics are described as:

$$\dot{v}_i = \psi(v_i) + I_i \tag{1}$$

$$I_i = W_i^E \emptyset(v^E) + W_i^r \emptyset(v^r) \tag{2}$$

,where $v = [v_1 . v_2 . \cdots . v_{n^r}] \in \mathbb{R}^{n^r}$ is the state or membrane potential of the RNN neurons, $\emptyset(\cdot)$ is a Lipschitz nonlinear dendrite such as $Sigmoid(\cdot).Tanh(\cdot)$, and, $\psi(v): \mathbb{R}^{n^r} \to \mathbb{R}^{n^r}$ is a Lipschitz leak-term where we used $-\alpha_l v$, $v^E, v^r$ in non-spiking neurons are directly equal to input $x$ and RNN state $v$, and in spiking neurons $v_i$ shows the filtered spike activity of neuron $i$. $S_i(t)$ is the spike train of the neuron $i$ and modelled as a sum of Dirac delta-functions:

$$q_i = \left(S_i \ * \kappa\right)(t) = \int_{-\infty}^{t} S_i(s)\kappa(t-s)ds \tag{3}$$

$$\kappa(t) = \exp(-t/\tau)/\tau \tag{4}$$

Moreover, a linear decoder is assumed as:

$$\hat{y} = h(v) = W^D v \tag{5}$$

## 1.2 Synaptic plasticity law

According to the equations mentioned in subsection 1.1, the RNN synaptic plasticity law can be modeled as a dynamical system. Consequently, principles of optimal control theory can be applied to derive a learning rule (optimal control law). For this purpose, we reformulate the AI tasks as a control problem. Thus, the output error were considered as follows:

$$e = \hat{y} - y \in \mathbb{R}^c \tag{6}$$

, where $c$ is a number of classes, $y \in \mathbb{R}^c$ denotes the desired output for the input pattern at time $t$, and $\hat{y} \in \mathbb{R}^c$ is the corresponding predicted output. By derivation of Eq. (6) with respect to $t$, the error dynamics can be described as:

$$\dot{e} = \dot{\hat{y}} - \dot{y} \ \in \mathbb{R}^c \tag{7}$$

Given that the decoder weights $W^D$ are assumed constant during the RNN training process, consequently, the classification error dynamics only depend on the RNN neural dynamics. Rewriting the neural space equations (1) and (2) in the form of Eq. (7) gives:

$$\begin{aligned} \dot{e} = W^D\dot{v} - \dot{y} &= W^D\left(\psi(v) + W^E\emptyset(v^E) + W^r\emptyset(v^r)\right) - \dot{y} \\ &= f(v) + g(v)u^r - f_d(v) - g_d(v)u_d^r \end{aligned} \tag{8}$$

neural space Eq. (8) can be considered as the following affine sate space equation from:

$$\dot{e} = f(e) + g(e)u \tag{9}$$

Where the control input is considered as part of parameters of the RNN, which the number is selected according to the problem and training data:

$$u^r = [u^f; u] = vec(W^r) \in \mathbb{R}^{N^r} \tag{10}$$

, in which $N^r = n^r \times n^r$ is the number of RNN parameters, $vec(\cdot)$ is vectored form of a matrix, $u^f \in \mathbb{R}^{N^r-N}$ is fixed RNN parameters, $u \in \mathbb{R}^N$ is the RNN plastic synapse or the control input and $[u^f; u]$ is combination of $u^f$and $u$. Therefore, we have the following nonlinear continuous-time equations:

$$\begin{aligned} f_1(v) &= W^D\left(W^E \emptyset(v^E) + \psi(v) + W^f \emptyset(v^f)\right) = f^E(v) + f^f(v) \\ f^E(v) &= W^D W^E \emptyset(v^E) \\ f^f(v) &= W^D(\psi(v) + W^f \emptyset(v^f)) \end{aligned} \tag{11}$$

$$g(v) = W^D \emptyset(v) \tag{12}$$

, where $f(e): \mathbb{R}^c \to \mathbb{R}^c$ is the drift dynamic of the system, and $g(e): \mathbb{R}^c \to \mathbb{R}^{c\times N}$ is the input dynamic of the system. A common and recommended action in this regard is to add noise into $u$ and $g(e)$ to realize the Persistence of Excitation (PE) condition [1] to collect sufficient information about the unknown parameters and make $g(v)$ full rank.

In all tasks of AI, the goal of learning is to adapt the RNN's parameters $W^r$ such that the error in each timestamp of the input pattern is minimized and the RNN's parameters or control input remains bounded. Thus, the following cost function was defined:

$$\min_{u(\cdot)\in U} \Im\big(u(\cdot).e(\cdot)\big) = \int_t^{t_f} \ell\big(e(\tau).u(\tau)\big)d\tau \tag{13}$$

$$\ell(e.u) = \eta e^T e + u^T R u \tag{14}$$

, wherein $\ell(\cdot.\cdot)$ is the utility function, $t_f$ is the last element of input patterns, $R$ symmetric positive definite matrix and $\eta$ is a constant positive hyper-parameter for ensuring that the error in cost function is sufficiently affective. Hence, the optimal value function can be written:

$$V(e) = \underset{u(\cdot)\in U}{Inf}\, \Im(u.e) \tag{15}$$

Under typical assumptions, $f(0) = 0$ and $f(v) + g(v)u$ is Lipschitz continues on a set $\Omega \in \mathbb{R}^n$ that contains the origin. Ultimately, it is desirable to achieve an optimal input control (weight update law) $u^*$ that stabilizes the system Eq. (11) and minimizes the cost function Eq. (15). This kind of input control $u$ is called admissible control [2].

Now, the learning rule has been formulated given the error dynamic in Eq. (11) and the cost function Eq. (15). To solve this dynamic optimization problem, the HJB equation is utilized, so the Hamiltonian of the cost function Eq. (15) associated with control input $u$ is defined as:

$$\mathcal{H}(e.u.V_e) = \ell(e.u) + V_e^T(\dot{e}) \tag{16}$$

, where $V_e = \partial V/\partial e$ is the partial derivative of the cost function, for admissible control policy $\mu$ we have:

$$\mathcal{H}(e.\mu.V_e) = 0 \tag{17}$$

The present study assumed that the solution to Eq. (17) is smooth giving the optimal cost function:

$$V^*(e) = \min_u\left(\int_t^{t_f} \ell\big(u(\tau).e(\tau)\big)d\tau\right) \tag{18}$$

, which satisfies the HJB equation

$$\min_u \mathcal{H}(e.u.V_e^*) = 0 \tag{19}$$

Assuming that the minimum on the left hand side Eq. (19) exists then by applying stationary condition $\partial\mathcal{H}(e.u.V_e)/\partial u = 0$, the learning rule (optimal control) can be obtained as:

$$u^*(e) = -\frac{1}{2}R^{-1}g^T(e)\,V_e^* \tag{20}$$

The optimal value function can be obtained as:

$$V^*(e) = \min_u(\int_t^{t_f} \eta e^T e + {u^*}^T R u^* d\tau) \tag{21}$$

Inserting this optimal learning rule Eq. (20) into nonlinear Lyapunov equation Eq.(16) gives the formulation of the HJB equation Eq. (19) in terms of $V_e^*$

$$0 = \eta e^T e + {V_e^*}^T(e) f(v) - \frac{1}{4}{V_e^*}^T(e) g(e) R^{-1} g^T(v) V_e^*(e) \tag{22}$$

Finding the learning rule for the RNN requires solving the HJB equation for the value function and then substituting the solution to obtain the desired learning rule. Although HJB gives the necessary and sufficient condition for optimality of a learning rule (control law) with respect to a loss function; Unfortunately, due to the nonlinear characteristics of the RNN, solving the HJB equation in explicit form is difficult or even impossible to derive for systems of interest in practice. Thus, the proposed framework focuses on the ADP method to approximate its solution. Thus, we will focus on the ADP method to approximate its solution. To address this issue, with the inspiration of the dopaminergic region structure and conformity with the Weierstrass high-order approximation theorem [3], we proposed a novel actor-critic based on the s-RC. Succinctly, the objective of tuning the critic weights is to minimize the Bellman equation error, and the objective of tuning the actor weights is to minimize the approximate value.

### *1.2.1 Critic network design*

In this subsection, a s-RC is exploited as critic NN to approximate the derivatives of the value function as follows:

$$\begin{aligned} &V_e\,(e) = W_c^D\, z_c + \varepsilon_c \in \mathbb{R}^{c\times 1} \\ &z_c \in \mathbb{R}^{n_c\times 1} \\ &W_c^D \in \mathbb{R}^{c\times n_c} \end{aligned} \tag{23}$$

Where $z_c$ is the new representation of input $e$, which is generated by the RNN and $W_c^D$ is the decoder weight matrix.

As far as the weights of critic decoder provide the best approximate solution for the Hamiltonian function Eq. (16) are unknown. So, it estimated with $\widehat{W}_c^D$:

$$\hat{V}_e(e) = \widehat{W}_c^D z_c \tag{24}$$

It is desired to select the weights of the critic network to minimize the loss function of the critic is defined as follows:

$$E_c(\widehat{W}_c^D) = \frac{1}{2} e_c^T e_c \tag{25}$$

Where for given any admissible policy $u$ the residual error:

$$\mathcal{H}(e.u.V_e) = \mathcal{H}(e.u.\widehat{W}_c^D) = \eta e^T e + u^T R u + (\widehat{W}_c^D z_c)^T(\dot{e}) = e_c \tag{26}$$

So, $\widehat{W}_c^D \to W_c^D$ and $e_c \to \varepsilon_c$.

The weight update law for the critic weights is gradient descent algorithm, that is:

$$\begin{aligned} \dot{\widehat{W}}_c^D &= -\alpha_c \frac{\partial E_c}{\partial \widehat{W}_c^D} = -\alpha_c (f(e) + g(e)u) z_c{}^T e_c \\ &= -\alpha_c \sigma_c (\sigma_c{}^T \widehat{W}_c^D + \eta e^T e + u^T R u) \end{aligned} \tag{27}$$

Where $\alpha_c > 0$ is learning rate and $\sigma_c = (f(e) + g(e)u) z_c{}^T$.

### *1.2.2 Actor network design*

Similar to the critic, another s-RC is used as the actor to approximate the synaptic plasticity rule (feedback control policy):

$$u(e) = W_a^D z_a + \varepsilon_a \in \mathbb{R}^{N \times 1} \tag{28}$$

$$z_a \in \mathbb{R}^{n_a \times 1}$$
$$W_a^D \in \mathbb{R}^{N \times n_a}$$

Let $z_a$ be the new representation of input $e$, $\widehat{W}_a^D$ an estimation of unknown matrix $W_a^D$ based on existing training data, so the feedback control policy can be expressed as:

$$\hat{u}(e) = \widehat{W}_a^D z_a \tag{29}$$

The loss function for the actor is defined:

$$E_a(\widehat{W}_a^D) = \frac{1}{2} e_a^T e_a \tag{30}$$

Where $e_a$ is define to be the difference between Eq. (29) and Eq. (20)

$$e_a = \hat{u} - u = \widehat{W}_a^D z_a + \frac{1}{2} R^{-1} g^T(e)(\widehat{W}_c^D z_c) \tag{31}$$

By applying the gradient descent, a weight update expression for the actor can be written as follows:

$$\dot{\widehat{W}}_a^D = -\alpha_a \frac{\partial E_a}{\partial \widehat{W}_a^D} = -\alpha_a (\widehat{W}_a^D z_a + \frac{1}{2} R^{-1} g^T(e)(\widehat{W}_c^D z_c)) z_a^T \tag{32}$$

Where $\alpha_a > 0$ is learning rate.

Finally, the RNN parameters update rule can be derived as follows:

$$vec(\widehat{W}^r) = \hat{u}(e) \tag{33}$$

This completes the RNN learning rule.

During the RNN learning process, the feedback pathway via actor-critic NNs forces the output of the network (estimated class) to follow the reference trajectory (true class), an effect that is widely used in control theory. After a sufficiently long learning time, the feedforward pathway without feedback can classify input patterns with acceptable accuracy.

## *1.3 The Analysis of stability and convergence*

In this subsection, the stability and convergence of the entire proposed framework and classification error will be analyzed. For this purpose, we need to show $\lim_{t\to\infty} ||\widetilde{W}_c^D|| = \lim_{n_e\to\infty} ||\widetilde{W}_c^D|| \to 0$, $\lim_{t\to\infty} ||\widetilde{W}_a^D|| = \lim_{n_e\to\infty} ||\widetilde{W}_a^D|| \to 0$ and $\lim_{t\to\infty} ||e|| = \lim_{n_e\to\infty} ||e|| \to 0$. In offline problems, $t = n_e n_l (n_s - 1) + i$ where $n_e$ is the number of the epochs, $n_s$ is the number of samples, $n_l$ is the time series length, and $i$ is the index of the current time series element. Since in offline problems the training dataset $n_s n_l$ is fixed so $t \to \infty$ as $n_e \to \infty$.

To prove asymptotically converge of the proposed framework, firstly, we need to show that the classification error $e$ and the RNN critic and actor decoder weight estimation errors $\widetilde{W}_c^D$, $\widetilde{W}_a^D$ are uniformly ultimately bounded (UUB). Therefore, we consider the following Lyapunov and, to facilitate the UUB analysis, the following assumption is made, which can reasonably be satisfied under the current problem settings.

**Assumption 1** Assume the following conditions:

a. We assumed $g(\cdot)$ is bounded to positive constants i.e., $\underline{g} \leq \|g(\chi)\| \leq \overline{g}$ therefore we have $\underline{G} \leq G \leq \overline{G}$ where $G = gR^{-1}g^T$.

b. The ideal weights of the critic and actor RNN decoder are upper bounded so that $\|W_c^D\| \leq \overline{W_c^D}$ and $\|W_a^D\| \leq \overline{W_a^D}$.

c. The approximation errors are upper bounded i.e., $\|\varepsilon_c\| \leq \overline{\varepsilon_c}$, $\|\varepsilon_a\| \leq \overline{\varepsilon_a}$.

d. The outputs of the RNNs critic, actor are bounded i.e., $\underline{z_c} \leq \|z_c\| \leq \overline{z_c}$, $\underline{z_a} \leq \|z_a\| \leq \overline{z_a}$.

According to PE assumption $\sigma_c = (f + gu){z_c}^T$ is bounded i.e., $\underline{\sigma_c} \leq ||\sigma_c|| \leq \overline{\sigma_c}$ .

Now, we consider the following Lyapunov function:

$$L(t) = L_{\widetilde{W}_c^D}(t) + L_{\widetilde{W}_a^D}(t) + L_e(t)$$

$$= \frac{1}{2\alpha_c} tr\left\{{\widetilde{W}_c^D}^T \widetilde{W}_c^D\right\} + \frac{\alpha_c}{2\alpha_a} tr\left\{{\widetilde{W}_a^D}^T \widetilde{W}_a^D\right\} + \alpha_c\alpha_a(e^Te + V(e)) \quad (34)$$

Where $\alpha > 0$, and $\widetilde{W}_c^D$ is the weight estimation error of the critic decoder, which is defined as:

$$\widetilde{W}_c^D = \widehat{W}_c^D - W_c^D \quad (35)$$

In addition, the dynamics of $\widetilde{W}_c^D$ is expressed as:

$$\dot{\widetilde{W}}_c^D = -\alpha_c(\sigma_c{\sigma_c}^T\widetilde{W}_c^D + \sigma_c\varepsilon_H) \quad (36)$$

Where

$$\varepsilon_H = -\varepsilon_c(f + gu) \tag{37}$$

Moreover, the actor weight estimation error is define as:

$$\widetilde{W}_a^D = \widehat{W}_a^D - W_a^D \tag{38}$$

Combining (23) and (28) with (32) yields:

$$\begin{aligned}\dot{\widetilde{W}}_a^D &= -\alpha_a\left(\widetilde{W}_a^D z_a + \varepsilon_a + \frac{1}{2}R^{-1}g^T(e)\left(\widetilde{W}_c^D z_c + \varepsilon_c\right)\right){z_a}^T \\ &= -\alpha_a\left(\widetilde{W}_a^D z_a + \frac{1}{2}R^{-1}g^T(v)\left(\widetilde{W}_c^D z_c\right) + \varepsilon'_a\right){z_a}^T\end{aligned} \tag{39}$$

Which $\varepsilon'_a$ becomes:

$$\varepsilon'_a = -(\varepsilon_a + \frac{1}{2}R^{-1}g^T(e)\varepsilon_c) \tag{40}$$

Thus, the time derivative of $L(t)$ is:

$$\begin{aligned}\dot{L}(t) &= \dot{L}_{\widetilde{W}_c^D}(t) + \dot{L}_{\widetilde{W}_a^D}(t) + \dot{L}_e(t) \\ &= \frac{1}{\alpha_c}tr\left\{\widetilde{W}_c^{D^T}\dot{\widetilde{W}}_c^D\right\} + \frac{\alpha_c}{\alpha_a}tr\left\{\widetilde{W}_a^{D^T}\dot{\widetilde{W}}_a^D\right\} + \alpha_c\alpha_a(2e^T\dot{e} \\ &+ \dot{V}(e))\end{aligned} \tag{41}$$

Substituting Eq. (36) into $\dot{\widetilde{W}}_c^D$ in Eq. (41), and assume $\left\|{\sigma_c}^T\widetilde{W}_c\right\| > \varepsilon_H$ the $\dot{L}_{\widetilde{W}_c^D}$ becomes:

$$\begin{aligned}\dot{L}_{\widetilde{W}_c^D}(t) &= \frac{1}{\alpha_c}tr\left\{\widetilde{W}_c^{D^T}\dot{\widetilde{W}}_c^D\right\} \\ &= -\frac{1}{\alpha_c}tr\left\{\alpha_c\widetilde{W}_c^{D^T}\sigma_c{\sigma_c}^T\widetilde{W}_c^D\right\} - \frac{1}{\alpha_c}tr\left\{\alpha_c\widetilde{W}_c^{D^T}\sigma_c\varepsilon_H\right\} \\ &\leq -(\underline{\sigma_c}^2)\left\|\widetilde{W}_c^D\right\|^2 + {\varepsilon_H}^2\end{aligned} \tag{42}$$

Substituting Eq. (39) into $\dot{\widetilde{W}}_a^D$ in Eq. (41), the second term becomes:

$$\begin{aligned}\dot{L}_{\widetilde{W}_a^D}(t) &= \frac{\alpha_c}{\alpha_a}tr\left\{\widetilde{W}_a^{D^T}\dot{\widetilde{W}}_a^D\right\} \\ &= -\frac{\alpha_c}{\alpha_a}tr\left\{\widetilde{W}_a^{D^T}\left[\alpha_a\left(\widetilde{W}_a^D z_a + \frac{1}{2}R^{-1}g^T(e)\left(\widetilde{W}_c^D z_c\right)\right.\right.\right. \\ &\left.\left.\left. + \varepsilon'_a\right){z_a}^T\right]\right\}\end{aligned} \tag{43}$$

According to Eq. (20), Eq. (29), and $\widetilde{W}_a^D z_a = -\frac{1}{2} R^{-1} g^T(e) \left( \widetilde{W}_c^D z_c \right)$ ,we have:

$$\dot{L}_{\widetilde{W}_a^D}(t) \leq -(\alpha_c \underline{z_a}^2) \left\| \widetilde{W}_a^D \right\|^2 + \frac{\alpha_c}{4\alpha_a} \|R^{-1}\|^2 \overline{g}^2 \overline{z_c}^2 \left\| \widetilde{W}_c^D \right\|^2 + \frac{\alpha_c}{\alpha_a} {\varepsilon'_a}^2 \tag{44}$$

Substituting Eq. (29) into the error dynamics Eq. (11), and based on Assumption 1 a., third term of Eq. (41) become:

$$\dot{L}_e(t) = \alpha_c \alpha_a (2e^T \dot{e} + \dot{V}(e)) = 2\alpha_c \alpha_a e^T \dot{e} + 2\alpha_c \alpha_a V_e^T \dot{e}$$

Substituting $\dot{e}$ with Eq. (9) and $V_e^T \dot{e}$ with $-(\eta e^T e + u^T R u)$ and according to Eq. (16) and Eq. (17):

$$\dot{L}_e(t) = 2\alpha_c \alpha_a e^T (f(e) + g(e) u) - \alpha_c \alpha_a (\eta e^T e + u^T u)$$

Substituting $u$ with Eq. (20) and Eq.(24) which will be equal to $u = -\frac{1}{2} R^{-1} g^T(v)\, \widehat{W}_c^D z_c$:

$$\begin{aligned} \dot{L}_e(t) &= 2\alpha_c \alpha_a e^T \left( f(e) - \frac{1}{2} g(e) R^{-1} g^T(e)\, \widehat{W}_c^D z_c \right) \\ &\quad - \alpha_c \alpha_a (\eta e^T e + u^T R u) \\ &= 2\alpha_c \alpha_a e^T \left( f(e) - \frac{1}{2} G \widehat{W}_c^D z_c \right) \\ &\quad - \alpha_c \alpha_a (\eta e^T e + u^T R u) \end{aligned} \tag{45}$$

In additional and according to Young inequality , we have:

$$e^T f(e) \leq \|e^T f(e)\| \leq \|e^T\| \|f(e)\| \leq \frac{1}{2} \|e\|^2 + \frac{1}{2} \|f(e)\|^2$$

Moreover, according to assumption 1 for $G \widehat{W}_c^D z_c$ we have:

$$
\begin{aligned}
\dot{L}_e(t) \le\ & \alpha_c\alpha_a\|e\|^2 + \alpha_c\alpha_a\|f(e)\|^2\|e\| + \alpha_c\alpha_a\overline{G}\overline{z_c}\overline{W_c^D}\ \|e\| \\
& - \alpha_c\alpha_a\eta\|e\|^2 - \alpha_c\alpha_a\lambda_{min}(R)\|u\|^2 \\
& \le -\alpha_c\alpha_a(\eta-1)\|e\|^2 \\
& + \alpha_c\alpha_a\left(\overline{G}\overline{z_c}\overline{W_c^D}\ + \|f(e)\|^2\right)\|e\| \\
& - \alpha_c\alpha_a\lambda_{min}(R)\|u\|^2
\end{aligned}
$$

Where $\lambda_{min}(R)$ is the minimum eigenvalue of $R$. Combining Eq. (42), Eq. (43) and Eq. (45):

$$
\begin{aligned}
\dot{L}(t) \le\ & -K_{\widetilde{W}_c^D{}^2}\left\|\widetilde{W}_c^D\right\|^2 - K_{\widetilde{W}_a^D{}^2}\left\|\widetilde{W}_a^D\right\|^2 - K_{u^2}\|u\|^2 - K_{e^2}\|e\|^2 + K_e\|e\| \\
& + K_1 \\
& = -K_{\widetilde{W}_c^D{}^2}\left\|\widetilde{W}_c^D\right\|^2 - K_{\widetilde{W}_a^D{}^2}\left\|\widetilde{W}_a^D\right\|^2 - K_{u^2}\|u\|^2 \\
& - K_{e^2}\left(\|e\| - \frac{K_e}{2K_{e^2}}\right)^2 + \frac{K_e{}^2}{4K_{e^2}} + K_1 \\
& \le -K_{\widetilde{W}_c^D{}^2}\left\|\widetilde{W}_c^D\right\|^2 - K_{\widetilde{W}_a^D{}^2}\left\|\widetilde{W}_a^D\right\|^2 - K_{u^2}\|u\|^2 \\
& - K_{e^2}\left(\|e\| - \frac{K_e}{2K_{e^2}}\right)^2 + \overline{K}
\end{aligned}
$$

*Where* (46)

$$K_{\widetilde{W}_c^D{}^2} = \underline{\sigma_c}^2 - \frac{\alpha_c}{4\alpha_a}\|R^{-1}\|^2\overline{g}^2\overline{z_c}^2$$

$$K_{\widetilde{W}_a^D{}^2} = \alpha_c\underline{z_a}^2$$

$$K_{e^2} = \alpha_c\alpha_a(\eta - 1)$$

$$K_e = \alpha_c\alpha_a\left(\overline{G}\overline{z_c}\overline{W_c^D}\ + \|f(e)\|^2\right)$$

$$K_{u^2} = \alpha_c\alpha_a\lambda_{min}(R)$$

$$K_1 = \varepsilon_H{}^2 + \frac{\varepsilon'_a{}^2}{2} \le \overline{K_1}$$

$$K = \frac{K_e{}^2}{4K_{e^2}} + K_1 \le \overline{K}$$

Thus, $\dot{L}(t) < 0$ if the proposed model hyper-parameters include $\alpha_c$. $\alpha_a$, $\eta$, R, $\psi(v)$ leaky term or $W^f \emptyset(q^f)$ fixed part of RNN's parameters in Eq. (11) are selected to satisfy three following inequalities:

$$0 < \alpha_c < \frac{4\alpha_a \underline{\sigma_c}^2}{\|R^{-1}\|^2 \overline{g}^2 \overline{z_c}^2} \tag{47}$$

And

$$0 < \alpha_a \tag{48}$$

And

$$\eta > 1 \tag{49}$$

Since all terms of $K$ are bounded consequently according to [4] the classification error, the critic and the actor errors are semi-globally uniformly ultimately bounded (SGUUB), which is absolutely enough to classification task. Nevertheless, for reaching more stable condition following inequalities could be hold:

$$\|e\| > \frac{K_e}{K_{e^2}} + \sqrt{\frac{K_1}{K_{e^2}}} \tag{50}$$

Or

$$\|\widetilde{W}_c^D\| > \sqrt{\frac{\overline{K}}{K_{\widetilde{W}_c}{}^2}} \tag{51}$$

Or

$$\|\widetilde{W}_a^D\| > \sqrt{\frac{\overline{K}}{K_{\widetilde{W}_a}{}^2}} \tag{52}$$

Or

$$\|u\| > \sqrt{\frac{\overline{K}}{K_u}} \tag{53}$$

Thus, according to the Lyapunov theory the classification error $e$, the RNN critic and actor decoder weight estimation errors $\widetilde{W}_c^D$, $\widetilde{W}_a^D$ and RNN parameters $u$ are UUB.

Our aforementioned analysis based on Lyapunov theory implies that $L_e(t)$ is bounded and decreasing. Therefore, we have:

$$L_e(t+T) = L_e(t) + \int_t^{t+T} \dot{L}_e(\tau) d\tau \leq \; L_e(t) - \gamma T \\ \Longrightarrow \; L_e(t) \leq L_e(t+T) \tag{54}$$

Where $\gamma > 0$. By substituting $T = n_l n_s$ which is training dataset length and according to the Lyapunov function properties [5] and summing both sides of Eq. (54) we have:

$$\sum_{t=0}^{n_l n_s} L_e(t) \leq \sum_{t=0}^{n_l n_s} L_e(t+T) \tag{55}$$

Consequently, Eq. (55) guarantee that the classification error in each epoch's is not more than the previous epochs. Thus, as the number of epochs increases, the classification training error tends to zero (i.e. $\lim_{t\to\infty} e(t) \to 0$).

In the same way, we can prove that for $L_{\widetilde{W}_c^D}(t)$ and $L_{\widetilde{W}_a^D}(t)$. This completes the stability and convergence proof.